\documentclass{article}

\usepackage[preprint]{neurips_2026}

\usepackage{amsmath,amsfonts}
\usepackage{algorithmic}
\usepackage{algorithm}
\usepackage{array}
\usepackage[caption=false,font=normalsize,labelfont=sf,textfont=sf]{subfig}
\usepackage{textcomp}
\usepackage{stfloats}
\usepackage{url}
\usepackage{verbatim}
\usepackage{graphicx}
\usepackage{cite}
\usepackage[utf8]{inputenc} 
\usepackage[T1]{fontenc}    
\usepackage{hyperref}       
\usepackage{booktabs}       
\usepackage{nicefrac}       
\usepackage{microtype}      
\usepackage{xcolor}         

\usepackage{amssymb}
\usepackage{mathtools}
\usepackage{amsthm}

\usepackage[capitalize,noabbrev]{cleveref}

\usepackage[textsize=tiny]{todonotes}

\usepackage{colortbl}

\usepackage{amsbsy}

\usepackage{bm}
\usepackage{adjustbox}
\usepackage{tabularx}
\usepackage{bbding}
\usepackage{transparent}
\usepackage{multirow}

\usepackage{wrapfig}

\title{Improving Full Waveform Inversion in Large Model Era}

\author{%
  Yinan Feng \\
  School of Data Science and Society\\
  University of North Carolina at Chapel Hill\\
  Chapel Hill, NC, USA \\
  \texttt{ynf@unc.edu} \\
  \And
  Peng Jin \\
  The Pennsylvania State University \\
  State College, PA, USA \\
  \texttt{pqj5125@psu.edu} \\
  \And
  Yuzhe Guo \\
  School of Data Science and Society\\
  University of North Carolina at Chapel Hill\\
  Chapel Hill, NC, USA \\
  \texttt{guoyuz@unc.edu} \\
  \AND
  Yinpeng Chen \\
  Google DeepMind \\
  \texttt{yinpengc@google.com} \\
  \And
  Youzuo Lin \\
  School of Data Science and Society\\
  University of North Carolina at Chapel Hill\\
  Chapel Hill, NC, USA \\
  \texttt{yzlin@unc.edu} \\
}

\begin{document}

\maketitle

\begin{abstract}
Full Waveform Inversion (FWI) is a highly nonlinear and ill-posed problem that aims to recover subsurface velocity maps from surface-recorded seismic waveforms data.
Existing data-driven FWI typically uses small models, as available datasets have limited volume, geological diversity, and spatial extent, leading to substantial concerns about overfitting. Although they perform well on synthetic datasets, current methods fail to generalize to more realistic geological structures.
In this work, we show that a model trained entirely on simulated and relatively simple data can generalize remarkably well to challenging and unseen geological benchmarks. We provide a working recipe that tames a billion-parameter model for FWI through coordinated scaling across three axes: model capacity, data diversity, and training strategy.
Our model achieves state-of-the-art performance on OpenFWI and significantly narrows the generalization gap in data-driven FWI. 
Across six challenging geophysical benchmarks, including Marmousi, 2D SEG/EAGE Salt and Overthrust, 2004 BP, Sigsbee, and SEAM Phase I, it infers complex structures absent from the training set and delivers significant performance improvements (SSIM from 0.5844 to 0.7669).
Overall, our results demonstrate that with an appropriate scaling strategy, large models trained on simple synthetic data can achieve substantial generalization to more complex and realistic geological structures.

\end{abstract}

\section{Introduction}
\label{sec:intro}

\begin{figure}[t]
    \centering
    \includegraphics[width=0.7\linewidth]{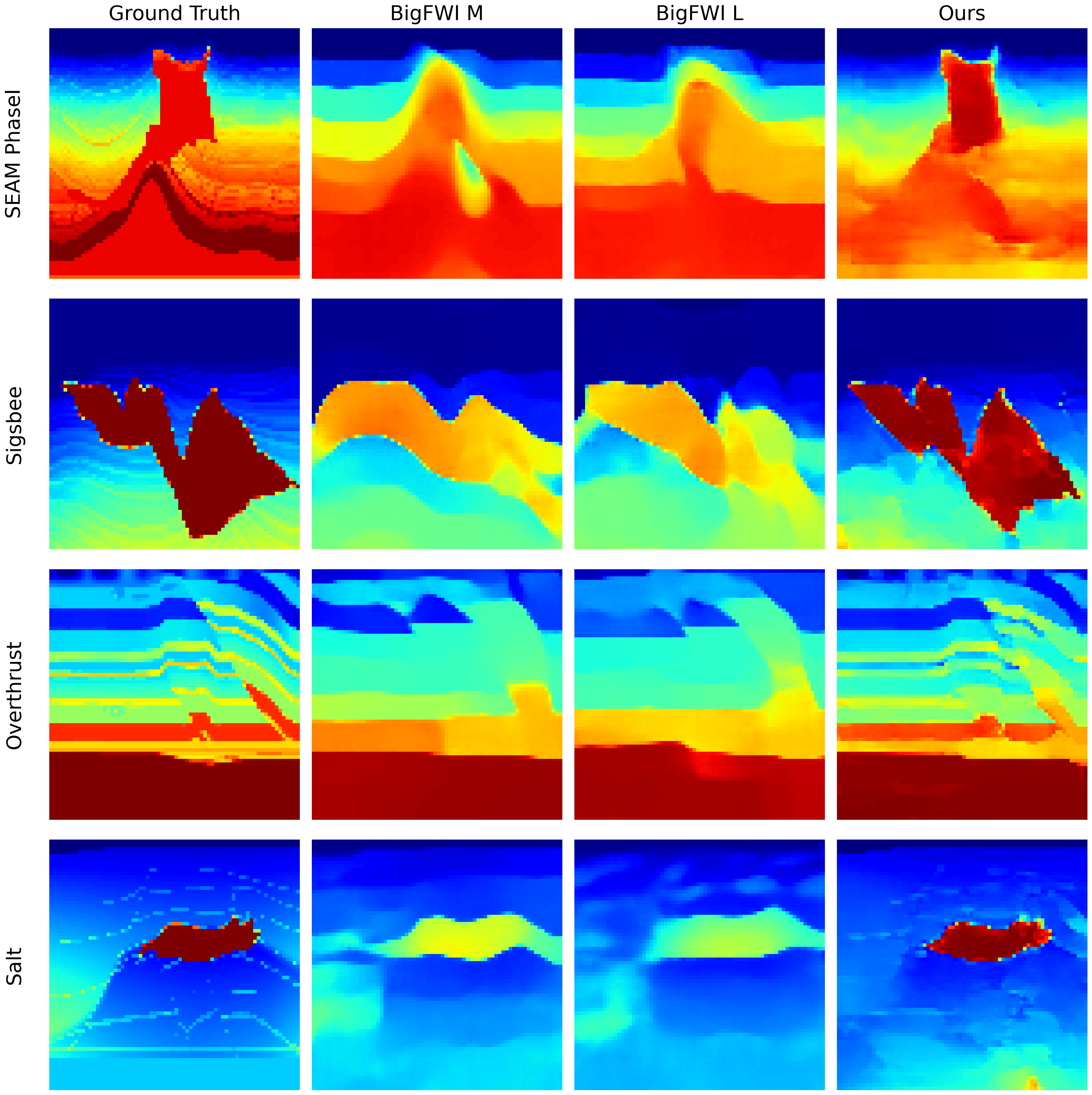}
    \caption{
    Reconstruction results on challenging realistic benchmarks. 
    BigFWI tends to collapse toward over-smoothed, mean-shaped solutions that miss key interfaces and salt bodies. 
    In contrast, our model produces sharper boundaries and geologically meaningful structures.
    }
    \vspace{-1mm}
    \label{fig:teaser_real}
    \vspace{-4mm}
\end{figure}

Subsurface imaging is critical in scientific and industrial applications, including earthquake monitoring \citep{virieux2017introduction,tromp2020seismic}, greenhouse gas storage \citep{li2021co2,wang2023default}, medical imaging \citep{guasch2020full,lozenski2024learned}, and oil and gas exploration \citep{virieux2009overview,wang2018microseismic}. At its core,  Full Waveform Inversion (FWI) serves as a representative method, which aims to reconstruct subsurface velocity maps from seismic data recorded on the surface, governed by the acoustic wave equation:
\begin{equation}
\label{eq:wave_eq}
\frac{\partial^2 p(x, z, t)}{\partial t^2} = v^2(x, z) \nabla^2 p(x, z, t) + f(x, z, t),
\end{equation}
where \( p(x, z, t) \) is the pressure wavefield at spatial location $(x, z)$ and time $t$, \( v(x, z) \) is the 2D vertical velocity map, \( \nabla^2 \) is the Laplacian operator, and \( s(x, z, t) \) is the source term. In practice, seismic data are often the waveform data collected at the surface (i.e., $p(x,z=0,t)$).
While FWI has the potential to produce high-resolution velocity maps, the inverse problem itself is inherently non-linear and ill-posed. In addition, the traditional iterative solvers based on gradient descent or adjoint-state methods are computationally demanding and sensitive to initialization, noise, and local minima \citep{virieux2009overview,pratt1999seismic,fichtner2010full}. These challenges have motivated growing interest in data-driven deep learning methods.


Recent advances in data-driven FWI have demonstrated the potential of deep neural networks to directly infer velocity maps from seismic data~\citep{araya2018deep,wu2019inversionnet,zhang2019velocitygan,li2020c,feng2022intriguing,jin2021unsupervised}. 
However, these models typically operate at small to medium model scales. This is largely due to the limited scale of existing FWI datasets in terms of data volume, geological diversity, and spatial extent, which raises strong concerns about overfitting. 
Prior studies also report that modest increases in model size do not reliably improve performance~\citep{jin2024empirical}.  
Although current approaches achieve promising results on synthetic datasets, they struggle to generalize to more realistic geological structures, particularly those containing complex features such as salt bodies or strong heterogeneity. 
As illustrated in the second and third columns of Fig.~\ref{fig:teaser_real}, a state-of-the-art end-to-end method, BigFWI~\citep{jin2024empirical}, often collapses into overly smoothed, mean-shaped solutions that fail to recover key interfaces or salt bodies, resulting in reconstructions that lack meaningful geological structure.

In this work, we show that a large model trained entirely on simulated and relatively simple data can generalize remarkably well to challenging and unseen geological benchmarks. However, deploying large models for a scientific inverse problem is far from straightforward. Several fundamental questions arise.
First, which architectural design can best capture seismic–velocity relationships, where accuracy and physical consistency are essential?
Second, how can one mitigate data scarcity when existing datasets are orders of magnitude smaller than what large models typically require?
Third, what learning strategies can effectively align the network with the underlying physical dependencies of wave propagation?
Finally, how can post-processing be leveraged to enhance physical consistency and further improve reconstruction accuracy?
We address these challenges by introducing a working recipe that tames a billion-parameter model for FWI through coordinated scaling across three axes: model capacity, data diversity, and training strategy. As illustrated in the rightmost column of Fig.~\ref{fig:teaser_real}, our method delivers significantly better reconstructions, recovering sharper layer boundaries and geologically meaningful high-velocity regions that prior methods fail to capture.

At the model level, our framework employs a one-billion-parameter transformer backbone that enables global contextual modeling between seismic and velocity tokens. 
Rather than relying on causal next-token generation, we adopt \emph{non-causal parallel decoding}, allowing all velocity tokens to be generated simultaneously with full self-attention, significantly improving accuracy and end-to-end efficiency by avoiding the autoregressive loop. 
Complementing this backbone, we employ a ViT-VQGAN tokenizer~\citep{yu2021vector} with an expanded bottleneck width that preserves fine geological details, avoiding the excessive information compression common in conventional VQGANs. 
This is particularly well-suited for scientific inversion tasks where spatial fidelity is critical.



To address data scarcity, we train a latent diffusion model~\citep{rombach2022high} on the \textsc{OpenFWI} dataset~\citep{deng2022openfwi} to synthesize additional velocity maps.
Each generated sample is paired with forward-simulated seismic data, ensuring physical consistency between the two modalities.
This process expands the training corpus from 408k to over five million velocity–seismic pairs and introduces hybrid geological structures that mix heterogeneous features across sub-datasets.
By substantially increasing both the quantity and structural diversity of training data, this diffusion-driven augmentation effectively meets the data requirements of the large transformer without relying on any real or realistic geological models.

\begin{wrapfigure}{r}{0.6\textwidth} 
  \vspace{-7mm}
  \centering
  \includegraphics[width=\linewidth]{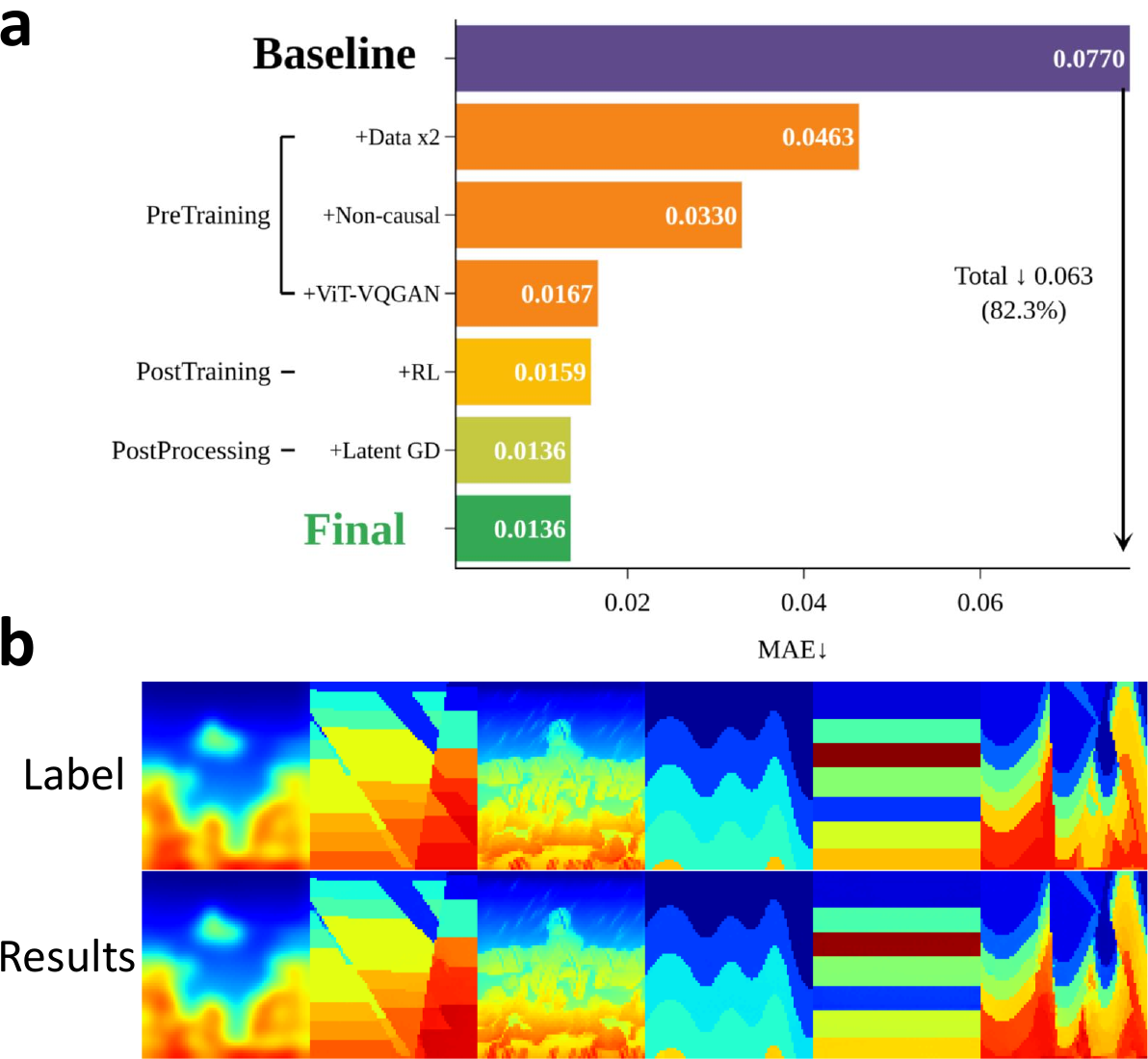}
  \vspace{-2mm}
    \caption{Overview of progressive model enhancements and final reconstruction quality. (a) Step-wise performance improvement as each component is introduced—from baseline to data augmentation, non-causal modeling, ViT VQGAN (without bottleneck), reinforcement learning alignment, and latent gradient refinement. (b) Illustration of the final reconstruction results, showing sharp structures and fine geological details across diverse scenarios.}
    \label{fig:teaser}
    \vspace{-5mm}
\end{wrapfigure}



In terms of learning strategies, we adopt a two-stage training pipeline.
In the \emph{supervised pre-training} stage, the model learns token-wise mappings from seismic-conditioned inputs to velocity representations.
In the \emph{RL-based post-training} stage, we further enhance model robustness through reinforcement learning (RL) optimization~\citep{deepseek-math,zheng2025group}.
We formulate velocity token generation as a discrete policy optimization problem, and utilize map-level rewards that encourage geological continuity and physical plausibility, enabling the model to move beyond the purely supervised learning.



In the post-processing stage, we first perform a physics-based gradient descent (GD) refinement to improve the physical consistency of the predicted velocity maps.
This step optimizes the continuous latent embeddings in the VQGAN decoder with respect to forward-modeling residuals, enforcing consistency with the governing equation.
To further mitigate the ill-posedness, we employ stochastic sampling and ensemble aggregation to combine multiple reconstructions when evaluating on realistic and unseen datasets, thereby reducing ambiguity and reducing predictive uncertainty.



Through extensive experiments, we demonstrate that each component of our large-model recipe contributes to improved performance and stability.
Figure~\ref{fig:teaser} summarizes the progressive gains and final reconstruction quality on \textsc{OpenFWI} achieved by different components.
Overall, our approach achieves state-of-the-art results on the synthetic \textsc{OpenFWI} benchmark and, more importantly, substantially narrows the long-standing generalization gap on realistic geophysical datasets, including Marmousi, 2D SEG/EAGE Salt and Overthrust, 2004 BP, Sigsbee, and SEAM Phase~I.

\section{Related Works}

Recently, deep learning–based FWI methods have achieved notable performance improvements, ranging from purely data-driven approaches to those that incorporate computationally intensive forward operators for enhanced physical consistency~\citep{Physics-2023-Lin,Deep-2021-Adler,Deep-2021-Yu}.
Fully supervised approaches~\citep{araya2018deep, wu2019inversionnet, zhang2019velocitygan, li2020c} learn direct mappings from seismic data to velocity maps using paired datasets. 
Feng et al.~\citep{feng2022intriguing, feng2024auto} decouple the seismic encoder and velocity decoder by leveraging latent-space correlations, enabling modular training. 
SiameseFWI~\citep{SiameseFWI-2024-Saad} adopts a Siamese architecture to align simulated and observed waveforms, facilitating cross-domain consistency. 
Unsupervised methods such as UPFWI~\citep{jin2021unsupervised} and Jia et al.~\citep{Seismic-2025-Jia} go further by removing the need for labels entirely. 
These approaches minimize waveform discrepancies under physical constraints using differentiable forward operators to optimize velocity predictions, though sacrificing accuracy and depending on computationally intensive solvers. 

Generative models have also been explored for FWI and velocity model building, either as priors for inverse recovery or as standalone generators of physically meaningful models. 
Diffusion-based priors can guide inverse recovery through plug-and-play (PnP) denoising~\citep{song2022solving, chung2023diffusion, zhang2025improving}, and Wang et al.~\citep{Prior-2023-Wang} show that learned diffusion priors can regularize the FWI optimization landscape. 
Beyond inverse recovery, WaveDiffusion~\citep{feng2024wavediffusion} jointly generates seismic data and velocity maps in a shared latent space, aiming to produce physically consistent paired samples. 
VelocityGPT~\citep{harsuko2025propagating, harsuko2025velocity} explores autoregressive transformer-based velocity generation from shallow to deep with uncertainty quantification, while incorporating multimodal priors such as well logs and RTM images to guide conditional sampling.

Large language models (LLMs) have revolutionized natural language processing (NLP), achieving state-of-the-art performance across a broad range of tasks, from text generation \citep{radford2018improving, brown2020language, jaech2024openai} to complex reasoning and mathematical problem solving \citep{guo2025deepseek}. 
Motivated by this success, recent research has sought to extend LLM principles to perception and multimodal domains, giving rise to Large Vision Models (LVMs) that sequentialize visual inputs~\citep{bai2024sequential} and unified multimodal architectures such as MMaDA~\citep{yang2025mmada} that align reasoning across text and image modalities. 
Despite this rapid progress, scientific inverse problems, particularly those governed by physical processes such as wave propagation, remain underexplored within this paradigm.

\section{Methodology}
\label{sec:method}


In this section, we outline the trajectory from a baseline toward a modern large-model FWI system based on discrete sequence modeling.
Figure~\ref{fig:teaser} summarizes the step-wise performance improvements contributed by each component and presents the final reconstruction results.
Our roadmap is as follows. We begin with an autoregressive baseline using two tokenization schemes, and then progressively introduce a sequence of design decisions:
(1) data scaling and augmentation,
(2) a non-causal transformer for parallel token prediction,
(3) a ViT-VQGAN tokenizer without a compression bottleneck, and post-training modules including
(4) policy optimization with velocity-level rewards (RL), and
(5) latent-space gradient refinement.
Together, these components yield state-of-the-art reconstruction accuracy and strong generalization ability.

\subsection{Baseline: Auto-regression Model}

\paragraph{Task formulation.}
Given seismic data $\mathbf{s}\in\mathbb{R}^{N_s\times T\times R}$, 
where $N_s$, $T$, and $R$ denote the number of sources, temporal steps, and receivers, respectively,
and its corresponding ground-truth velocity map $\mathbf{v}\in\mathbb{R}^{H\times W}$, the goal of the backbone model is to predict the discretized velocity token sequence $\mathbf{y}=(y_1,\dots,y_{L_v})$ whose indices correspond to entries in a pretrained VQGAN codebook.
The conditional generation process follows an autoregressive formulation:
\begin{equation}
p_\theta(\mathbf{y}\mid \mathbf{s})
= \prod_{i=1}^{L_v} p_\theta\!\left(y_i \mid \mathbf{s}, y_{<i}\right),
\end{equation}
where $\theta$ are backbone model parameters. 
During inference, the predicted sequence $\mathbf{y}$ is decoded by the VQGAN decoder
to reconstruct the full-resolution velocity map $\hat{\mathbf{v}}$.

The auto-regressive baseline comprises the following pipeline:
(1) seismic data are tokenized via patch embedding;
(2) velocity maps are quantized by a frozen VQGAN encoder;
(3) both are concatenated into a unified token sequence;
(4) a causal transformer predicts velocity tokens autoregressively; and
(5) predicted tokens are decoded into inversion result $\hat{\mathbf{v}}$.

\paragraph{Tokenization.}
The model employs two independent tokenizers that embed seismic and velocity data
into their corresponding token spaces.

\textbf{(a) Seismic tokenizer.}
The seismic data is first divided into spatio-temporal patches and projected through a linear embedding layer (\texttt{PatchEmbed}) to form a continuous token sequence  
$\mathbf{m}\in\mathbb{R}^{L_s\times C_h}$.
A positional encoding $\mathbf{E}_{\text{pos}}^{(s)}$ is added to preserve ordering:
\begin{equation}
\mathbf{m} = \text{PatchEmbed}(\mathbf{s}) + \mathbf{E}_{\text{pos}}^{(s)}. \label{s_emb}
\end{equation}

\textbf{(b) Velocity tokenizer.}
A pretrained VQGAN encoder discretizes the velocity map $\mathbf{v}$ into a latent grid 
and quantizes it into codebook tokens. 
Each latent feature is linearly projected to match the backbone model’s hidden dimension 
and combined with positional embeddings:
\begin{equation}
\mathbf{t} = \text{Linear}\!\left(\text{Enc}(\mathbf{v})\right)
+ \mathbf{E}_{\text{pos}}^{(v)}.
\end{equation}
While the linear projection is trained together with the large transformer model, the VQGAN tokenizer remains frozen to ensure a consistent latent representation 
capturing fine geological structure.

\paragraph{Input Sequence construction.}
We concatenate the seismic and velocity embeddings 
into a single multimodal ``visual sentence'':
\begin{equation}
[\text{BOS},\langle B_{\text{seis}}\rangle,\mathbf{m},\langle E_{\text{seis}}\rangle,
\langle D_{\text{id}}\rangle,\langle B_{\text{vel}}\rangle,\mathbf{t}],
\end{equation}
where special tokens
\texttt{BOS} marks the beginning of the sequence;
$\langle B_{\text{seis}}\rangle$ and $\langle E_{\text{seis}}\rangle$ delimit the start and end of the seismic token segment $\mathbf{m}$;
$\langle D_{\text{id}}\rangle$ is an auxiliary task-specific identifier token that is trained to encode the inversion objective or dataset identity to help model learning;
$\langle B_{\text{vel}}\rangle$ denotes the beginning of the velocity segment;
and $\mathbf{t}$ represents the ground-truth velocity tokens from the VQGAN tokenizer.

\paragraph{Network backbone and causal decoding.}
The backbone is a causal transformer with 1B parameters. 
During training, only the velocity positions are supervised with the standard token-level cross-entropy loss, while all preceding tokens serve as conditioning context.
In inference, the transformer generates velocity tokens sequentially:
\begin{equation}
y_i' = \arg\max_{y} p_\theta(y \mid \mathbf{s}, y_{<i}), \quad i=1,\dots,L_v.
\end{equation}
The completed sequence $\mathbf{y}'$ is then decoded through the VQGAN' decoder 
to produce $\hat{\mathbf{v}}$:
\begin{equation}
\hat{\mathbf{v}} = \text{Dec}(\mathbf{y}').
\end{equation}

While this formulation enables explicit sequence modeling of seismic-velocity relations, it enforces strictly sequential decoding and limits global token interaction.
These limitations motivate the transition to non-causal, parallel predictions.

\subsection{Macro Design}

\paragraph{Data Augmentation.}
To enhance data diversity and mitigate overfitting, we construct a large-scale synthetic dataset through a two-stage generation pipeline. 
First, a latent diffusion model is trained in the velocity domain to synthesize subsurface structures, enabling sampling of geologically diverse velocity maps.
Second, for each synthesized velocity map $\tilde{\mathbf{v}}$, an acoustic forward simulator generates the corresponding seismic observation $\tilde{\mathbf{s}} = f(\tilde{\mathbf{v}})$, ensuring physical consistency between the two modalities.
This process enlarges the training corpus by approximately tenfold beyond the available paired samples, encompassing a wider range of geophysical conditions and acquisition geometries, which introduces more diverse synthetic cases that mix heterogeneous geological features from different subdatasets in the original dataset.

\paragraph{Non-Causal Sequence Modeling.}
While the baseline autoregressive transformer predicts velocity tokens sequentially, we reformulate inversion as a \emph{non-causal, parallel prediction} problem to exploit bidirectional context fully.
The model operates on a unified input sequence constructed as
\begin{equation}
[\text{BOS},\langle B_{\text{seis}}\rangle,\mathbf{m},\langle E_{\text{seis}}\rangle,
\langle D_{\text{id}}\rangle,\langle B_{\text{vel}}\rangle,\mathbf{p}_{\text{vel}}],
\end{equation}
where $\mathbf{p}_{\text{vel}}$ denotes learnable placeholder tokens corresponding to velocity-token positions.
These placeholders are optimized jointly with the transformer to represent the output embedding space. 
Unlike causal decoding, all velocity tokens are generated simultaneously, with the attention mask set to permit full self-attention across seismic and velocity segments. 
This non-causal formulation enables global token interaction, improving reconstruction accuracy and providing substantially better end-to-end efficiency in our fixed-token setting.

\paragraph{ViT-VQGAN Tokenizer.}
To obtain discrete and high-fidelity representations of velocity maps, we replace the original pretrained VQGAN with a ViT-based architecture~\citep{yu2021vector}.
Unlike the previous architecture, this variant removes the compression bottleneck by interpolating the input and output to a resolution five times larger, enabling the preservation of fine-scale geological structure.
Rotary positional embeddings (RoPE)~\citep{su2021roformer} are applied within attention layers to maintain spatial coherence.
This tokenizer achieves substantially lower reconstruction error with a smaller network, providing an effective discrete representation space for large-scale sequence modeling.

\subsection{Post-Training Enhancements}
\label{sec:post_training}

The macro design establishes a scalable sequence-modeling framework for FWI.
To further enhance physical fidelity and convergence, we introduce two complementary post-training strategies that strengthen the model from different perspectives: a reinforcement learning (RL) stage that aligns model outputs with velocity-level rewards, and a latent-space gradient refinement mechanism that corrects residual artifacts in the decoder’s continuous representation. 
Each stage builds upon the trained model without re-training the tokenizer, progressively improving the reconstruction quality and physical consistency.

\paragraph{Policy Optimization with Velocity-Level Rewards.}
To further align model outputs with high-level structural fidelity, we employ a reinforcement-learning-style fine-tuning procedure that treats the trained transformer as a stochastic policy $\pi_\theta(\mathbf{y}\mid \mathbf{s})=
\prod_{i=1}^{L_v}p_{\theta}(y_i|\mathbf{s})$. Velocity tokens are sampled independently across spatial positions.
The policy is optimized to maximize the expected reward associated with the generated velocity map:
\begin{equation}
\max_{\theta} \; \mathbb{E}_{\mathbf{y}\sim\pi_\theta(\cdot\mid\mathbf{s})}\!\left[ -\|\mathbf{v}-\hat{\mathbf{v}}(\mathbf{y})\| \right].
\end{equation}
Then a GRPO-style optimization \citep{zheng2025group} was applied. This contrasts with purely supervised cross-entropy training and encourages the model to generate token sequences that are holistically better at the map level, rather than merely correct on a per-token basis.
The RL stage further reweights the model’s token predictions to prefer sequences that produce geologically consistent velocity fields, serving as an improvement mechanism that bridges supervised objectives with structural priors.





\paragraph{Latent-Space Gradient Refinement.}
Finally, we perform gradient descent (GD) refinement directly in the decoder’s latent space to correct residual local inconsistencies after token prediction.
Let $\mathbf{z}$ denote the continuous latent features produced by the VQGAN encoder prior to quantization, and let $\mathbf{z}_{\text{pred}}$ be the latent reconstructed from the predicted tokens.
Instead of updating the velocity map $\hat{\mathbf{v}}$ itself, as is done in traditional gradient-based FWI. we refine $\mathbf{z}_{\text{pred}}$ by optimizing a differentiable reconstruction objective:
\begin{equation}
\mathbf{z}_{k+1} = \mathbf{z}_k - \eta\,\nabla_{\mathbf{z}}\mathcal{L}_{\mathrm{rec}}\!\left(\mathcal{F}(\text{Dec}(\mathbf{z}_k)), \mathbf{s}\right),
\end{equation}
where $\eta$ is the learning rate and $\mathcal{L}_{\mathrm{rec}}$ is a reconstruction loss on seismic data.
Because gradients are computed with respect to $\mathbf{z}$ rather than $\hat{\mathbf{v}}$, this refinement operates in a smooth latent manifold. 
This design differs from traditional FWI, which directly updates the velocity map and therefore requires strong regularization to avoid unstable or physically implausible solutions.
By refining in the latent space of the VQGAN, the model retains high-frequency geological details while improving global consistency, resulting in improved accuracy. 



\section{Experiments}
In this section, we systematically evaluate the proposed large-model FWI framework. We first describe the experimental setup, including datasets, preprocessing, and training configurations. We then present quantitative and qualitative results. Finally, we assess performance on more realistic and previously unseen geological data, demonstrating strong generalization beyond the training distribution. We provide additional ablations and analyses on reinforcement learning, tokenizer design, and the effects of model capacity and data scaling in Appendix~\ref{app:RL},~\ref{app:vq}, and~\ref{app:ablation}.

\subsection{Dataset and Training Setup}
\label{subsec:dataset_and_training}
In the experiments, we evaluate the performance of our model using the \textsc{OpenFWI} dataset \citep{deng2022openfwi}, a comprehensive benchmark collection comprising 11 2D subsets of realistic synthetic seismic data paired with subsurface velocity maps, designed explicitly for FWI tasks. These subsets represent diverse geological structures, including curved velocity layers, flat velocity layers, and flat layers intersected by faults. Our experiments utilized 10 subsets (Fault, Vel, and Style Families) with the same resolution to ensure a thorough assessment. There are a total of 408K training samples. For additional details, we refer readers to the original OpenFWI paper~\citep{deng2022openfwi}.


 \paragraph{Data Augmentation.}
To increase data diversity and meet the scale required by large-model training, we employ a generative augmentation strategy based on a latent diffusion model~\citep{rombach2022high} trained on the \textsc{OpenFWI} training set.
After training, synthetic velocity maps are sampled from the learned latent distribution, and each map is paired with its corresponding seismic data generated via the acoustic forward modeling operator.
The synthesized samples achieve a Fréchet Inception Distance (FID) of 260.33 when evaluated against the original \textsc{OpenFWI} dataset.
This process expands the number of training pairs from 408{,}000 to approximately 5 million, significantly improving data coverage and supporting large-scale model training.

\paragraph{Training Configuration.}
The VQGANs and latent diffusion model are trained on NVIDIA H100 GPUs, while the transformer backbone is trained on four NVIDIA L40 GPUs with a batch size of 8.
We use the AdamW optimizer with a learning rate of $5\text{e-}4$, weight decay of $5\text{e-}2$, and a cosine learning rate schedule with a 10-epoch warmup.
Training spans 100 epochs for the base models, with about 6 hours per epoch, and 700K iterations for reinforcement-learning fine-tuning.
All predicted velocity maps are evaluated using MAE under the normalized range $\left[-1, 1\right]$, consistent with prior work~\citep{wu2019inversionnet,feng2024auto,deng2022openfwi}.



\begin{wraptable}{r}{0.5\textwidth}
\vspace{-7mm}
\centering
\small
\caption{\textbf{Quantitative comparison with baseline methods,} on OpenFWI testset. Models are organized incrementally, with each row cumulatively including components introduced in prior rows.}
\label{tab:main_results}
\begin{tabular}{clcc}
\hline
\multicolumn{2}{c}{Model}               & MAE $\downarrow$ & \#Param \\ \hline
\multicolumn{2}{c}{BigFWI-M~\citep{jin2024empirical}}    & 0.0437           & 28M          \\
\multicolumn{2}{c}{BigFWI-L~\citep{jin2024empirical}}    & 0.0432           & 87M          \\ \hline
\multirow{3}{*}{\textbf{Causal}}       & Baseline                          & 0.0770           & 1.09B        \\
                                       & + Data x2                         & 0.0463           & 1.09B        \\
                                       & + Data x10                        & 0.0368           & 1.09B        \\\hline
\multirow{5}{*}{\textbf{Non-Causal}}   & Data x2                         & 0.0330           & 1.09B        \\
                                       & + ViT-VQGAN             & 0.0167           & 1.03B        \\
                                       & + RL       & 0.0159           & 1.03B        \\
                                       & + Latent GD   & \textbf{0.0136}  & 1.03B        \\ \hline 
\end{tabular}
\vspace{-8mm}
\end{wraptable}

\subsection{Comparison with Baseline Methods.}
Table~\ref{tab:main_results} reports the quantitative comparison between our model variants and the CNN-based BigFWI baselines~\citep{jin2024empirical}.
Across all settings, our framework significantly outperforms prior methods and demonstrates clear benefits from scaling data, architecture, and training strategy.

The causal baseline achieves only moderate accuracy (MAE~0.0770).
Due to the large model size and the limited scale of the \textsc{OpenFWI} training set, this causal baseline cannot operate at its full capacity, leading to suboptimal performance. Increasing the training data by a factor of two reduces the error to 0.0463, matching the performance range of BigFWI. Scaling to ten times more data further improves accuracy to 0.0368, indicating strong data-driven scalability. Note that, unlike the other variants, this model with 10 times data is trained for only 20 epochs with a 1-epoch warmup.

Switching from causal to non-causal decoding yields a larger improvement, reducing the MAE from 0.0463 to 0.0330 (under the same training data level) by allowing full bidirectional attention among all tokens.
Replacing the original U-Net VQGAN tokenizer with the ViT-VQGAN, which is trained on the full ×10 augmented dataset, leads to a substantial improvement (MAE~0.0167).
Notably, the final backbone transformer is trained only on the ×2 dataset. With the new ViT-VQGAN tokenizer and non-causal setting, increasing the data scale to ×10 under the same compute budget did not improve performance proportionally; ×10 data for 10 epochs achieved an MAE of 0.0287, while ×2 data for 50 epochs achieved 0.0247. We therefore use ×2 data in the final setting, suggesting that the main gains come from the improved tokenizer’s high-fidelity, non-compressed latent representation.
Reinforcement learning (RL) applied on top of the ViT-VQGAN version further reduces the error to 0.0159.
Finally, applying physics-based latent gradient descent (GD) refinement yields the best performance (MAE~0.0136), establishing a new state of the art in data-driven FWI.

\begin{wrapfigure}{r}{0.49\textwidth} 
  \vspace{-6mm}
  \includegraphics[width=0.95\linewidth]{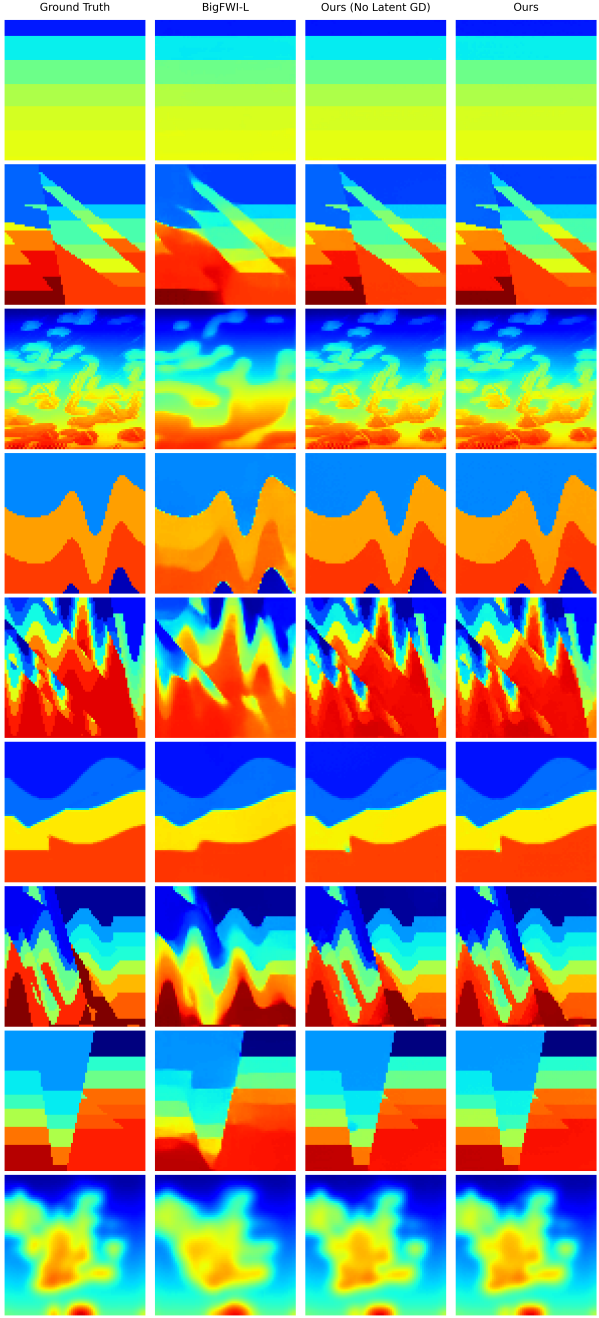}
\vspace{-2mm}
\caption{
\textbf{Qualitative comparison} of reconstructed velocity maps.}
\vspace{-5mm}
\label{fig:qualitative}
  \vspace{-18mm}
\end{wrapfigure}

Remarkably, our model reaches state-of-the-art end-to-end performance for data-driven FWI, even before any post-processing refinement is applied. Overall, these results demonstrate that data scaling, non-causal modeling, improved tokenization, RL alignment, and physics-based refinement collectively define a powerful and extensible recipe for large-model FWI, pushing performance well beyond existing baselines.

\paragraph{Qualitative Results.}
Figure~\ref{fig:qualitative} shows representative reconstructions from BigFWI-L, our model without latent GD refinement, and our full model. 
BigFWI-L produces overly smooth velocity maps that miss key structural boundaries, whereas our end-to-end model (without GD) already recovers sharp interfaces, coherent stratigraphy, and realistic fault geometries across diverse scenarios.

The addition of latent GD refinement yields less visible changes, as seen in the comparison between the “Ours (No Latent GD)” and “Ours” columns. Thus, the primary role of GD to remove small high-frequency artifacts that are imperceptible to the human eye, such as layer-wise inconsistencies or localized oscillations introduced by the network. 
By enforcing physical consistency, GD refinement smooths out these physically implausible variations, improving numerical accuracy and physical correctness without significantly changing the visual appearance.

\subsection{Zero-Shot Generalization on Realistic Geophysical Benchmarks}


To evaluate generalization beyond the in-distributions synthetic data used during training, we evaluate our model in a strict zero-shot setting on a collection of six widely used geophysical benchmark velocity models, including Marmousi~\citep{brougois1990marmousi}, 2D SEG/EAGE Salt and Overthrust~\citep{aminzadeh19963}, 2004 BP~\citep{billette20052004}, Sigsbee~\citep{paffenholz2002subsalt}, and SEAM Phase~I~\citep{fehler2011seam}.  
These datasets contain substantially more complex and heterogeneous geological structures than those seen in the \textsc{OpenFWI} training distribution. A prominent example is the presence of large salt bodies: high-velocity, sharply bounded formations that generate strong reflections and multipathing effects. Such structures create discontinuous wavefronts and non-convex inversion landscapes, making them uniquely challenging for FWI~\citep{virieux2009overview}.  
Their absence in the training data makes zero-shot reconstruction particularly difficult for data-driven approaches.

In particular, the Salt and Overthrust originate from 3D velocity volumes; we uniformly extract ten 2D slices from each volume. For SEAM Phase~I, we only use the $V_p$ as we use an acoustic forward modeling. All velocity maps are downsampled to $70 \times 70$ and linearly mapped to the range 1500–4500~m/s to match the normalization used during training. Seismic data are generated with the same acoustic forward-modeling operator as in \textsc{OpenFWI}: five evenly spaced sources with a 15~Hz Ricker wavelet, recorded by 70 receivers sampling 1,000 timesteps over one second.

\begin{wrapfigure}{r}{0.49\textwidth} 
  \vspace{-6mm}
  \centering
\includegraphics[width=\linewidth]{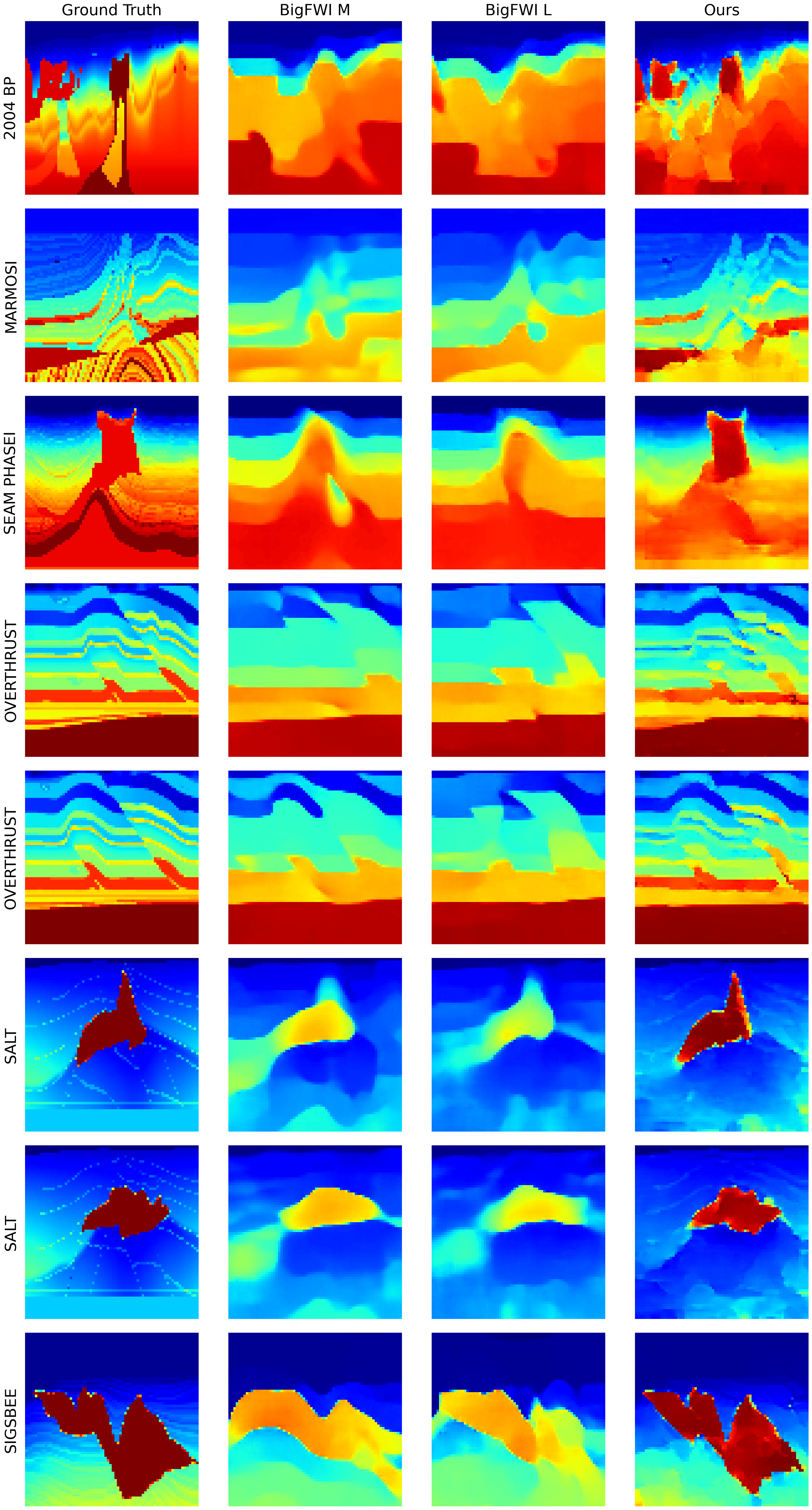}
\caption{
\textbf{Qualitative comparison of reconstructed velocity maps.}
Our method produces sharper and finer structures.}
\label{fig:vis_zero_shot}
\vspace{-9mm}
\end{wrapfigure}

For evaluation, our model predicts five samples per velocity map via stochastic sampling.  
Each sample is refined through latent space GD, and the final prediction is obtained by averaging the refined outputs.  We evaluate predicted velocity maps using MAE, RMSE, and Structural Similarity (SSIM). MAE and RMSE quantify pixel-wise errors, while SSIM captures perceptual similarity, reflecting the structured information of velocity maps where distortions can be easily perceived by a human. MAE and MSE are computed in the range $\left[1500, 4500\right]$, and SSIM in $\left[0, 1\right]$.

Quantitative results are reported in Table~\ref{tab:real_results}. The results for the Salt and Overthrust datasets are averaged over ten uniformly sampled 2D slices extracted from their 3D volumes. Across all six benchmark families and 26 velocity maps, our method achieves the best overall performance, producing substantially lower error and much higher SSIM than the BigFWI baselines. For ablation, without latent GD refinement, the ensemble prediction reaches an MAE of 148.35, while incorporating GD further reduces the MAE to 139.81.

While BigFWI attains better MAE in a few individual cases (e.g., SEAM Phase I and Salt), the corresponding reconstructions are severely oversmoothed and lack detailed structure. This is clearly visible in Figure~\ref{fig:vis_zero_shot}, where BigFWI tends to collapse toward mean-shaped solutions that fail to recover key interfaces or high-contrast anomalies. In contrast, our model reconstructs coherent stratigraphy, sharper layer boundaries, and geologically meaningful high-velocity regions.

These differences are particularly pronounced for salt-body targets. Despite having a large distribution shift, our model successfully delineates salt geometry and surrounding stratification, demonstrating strong generalization.
Overall, the zero-shot results show that our large-model paradigm not only reduces numerical error but also produces physically meaningful and interpretable reconstructions on challenging and previously unseen geological structures.

\begin{table*}[t]
\centering
\small
\setlength{\tabcolsep}{4pt}
\renewcommand{\arraystretch}{0.8}
\caption{\textbf{Quantitative performance across six benchmark families.}
Our method achieves substantially lower MAE/RMSE and higher SSIM, with especially strong gains on SSIM.}
\begin{tabular}{c|l|c|c|c|c|c|c|c}
\toprule
\multicolumn{1}{l|}{\textbf{Metrics}} &
\textbf{Model} &
\textbf{Overall} &
\textbf{2004 BP} &
\textbf{Marmosi} &
\textbf{SEAM Phase I} &
\textbf{Overthrust} &
\textbf{Salt} &
\textbf{Sigsbee}  \\
\hline
\multirow{3}{*}{MAE$\downarrow$} 
& BigFWI-M 
& 199.20 & 247.15 & 226.63 & \textbf{185.02} & 221.57 & \textbf{150.13} & 431.60 \\
& BigFWI-L
& 210.99 & 238.31 & 233.22 & 190.84 & 243.25 & 159.07 & {398.08} \\
& Ours
& \textbf{139.81} & \textbf{133.45} & \textbf{145.62} & {199.44} & \textbf{118.25} & {155.22} & \textbf{148.40} \\
\hline

\multirow{3}{*}{RMSE$\downarrow$} 
& BigFWI-M 
& 312.15 & 390.30 & 346.13 & 289.37 & 297.96 & {281.57} & 715.32 \\
& BigFWI-L
& 327.79 & 377.88 & 349.42 & 294.67 & 320.46 & 301.74 & {656.29} \\
& Ours
& \textbf{241.78} & \textbf{247.61} & \textbf{251.12} & \textbf{281.57} & \textbf{199.76} & \textbf{268.67} & \textbf{353.37} \\
\hline

\multirow{3}{*}{SSIM$\uparrow$} 
& BigFWI-M 
& 0.5844 & 0.5828 & 0.5074 & 0.6371 & 0.5020 & 0.6691 & 0.5852 \\
& BigFWI\_L
& 0.5579 & {0.5962} & {0.4930} & {0.6425} & 0.4516 & 0.6554 & {0.5973} \\
& Ours
& \textbf{0.7669} & \textbf{0.7560} & \textbf{0.7201} & \textbf{0.7804} & \textbf{0.7916} & \textbf{0.7425} & \textbf{0.8071} \\
\bottomrule

\end{tabular}
\label{tab:real_results}
\vspace{-6mm}
\end{table*}

\section{Conclusion}
We presented a large-model recept for data-driven Full Waveform Inversion that combines a billion-parameter non-causal transformer, a high-fidelity ViT-VQGAN tokenizer trained on diffusion-augmented data, RL-based post-training, and physics-guided latent refinement. This framework achieves a significant performance improvement on \textsc{OpenFWI} and shows strong generalization to more realistic and previously unseen geological structures. Although demonstrated under simplified 2D acoustic settings, our results provide clear evidence that scaling model capacity, data diversity, and physics alignment is a promising direction for seismic inversion. We expect future extensions toward elastic and 3D physics, variable survey geometries, and higher-resolution domains to further advance the frontier of large-model geophysical imaging.

\section{Acknowledgement}
Y. Lin was supported by a faculty start-up grant from the University of North Carolina at Chapel Hill School of Data Science and Society. This work was also supported by the U.S. Department of Energy Office of Science Advanced Scientific Computing Research program under Award No. DE-SC0025377 and by the U.S. National Science Foundation under Award No. 2504439.

{
\small
\bibliographystyle{ieeenat_fullname}
\bibliography{main}
}


\appendix

\section{Effectiveness of Reinforcement Learning} \label{app:RL}

To evaluate the effectiveness of reinforcement learning (RL), we apply RL to three representative base models with different configurations. The results are summarized in Table~\ref{tab:rl_effect}. RL consistently improves reconstruction accuracy across all settings.


For the causal transformer trained with doubled data (\textit{Causal + Data,x2}), RL yields a substantial 25.7\% reduction in MAE, demonstrating that policy optimization provides meaningful gains even under strictly autoregressive decoding.
A similar, though smaller, improvement is observed for the non-causal model (\textit{Non-Causal + Data,x2}), where RL reduces the MAE by 10.6\%, indicating that RL remains beneficial even when the model already leverages full bidirectional context during decoding.
RL also improves performance in our final ViT-VQGAN–enhanced model (\textit{Non-Causal + Data,x2 + ViT-VQGAN}).
Starting from an MAE of 0.0167, RL reduces the error to 0.0159, corresponding to a 4.7\% improvement.
As expected, the gains become smaller as the base model becomes stronger and harder to improve.
Overall, these results demonstrate that RL consistently yields measurable improvements across a broad range of model configurations.

\begin{table}[h]
\centering
\small
\caption{\textbf{Qualitative comparison of reinforcement learning} across different model configurations.}
\vspace{-2mm}
\label{tab:rl_effect}
\begin{tabular}{lc}
\toprule
\textbf{Model Configuration} & \textbf{MAE} $\downarrow$ \\
\midrule
Causal + Data x2 & 0.0463 \\
Causal + Data x2 + RL & 0.0344 \\
\midrule
Non-Causal + Data x2 & 0.0330 \\
Non-Causal + Data x2 + RL & 0.0295 \\
\midrule
Non-Causal + Data x2 + ViT-VQGAN & 0.0167 \\
Non-Causal + Data x2 + ViT-VQGAN + RL & \textbf{0.0159} \\
\bottomrule
\end{tabular}
\end{table}


\section{Comparison of VQGAN Tokenizers} \label{app:vq}
To evaluate reconstruction quality, we compare the conventional CNN-based VQGAN with the new ViT-based variant.
As shown in Table~\ref{tab:vqgan}, the ViT-VQGAN achieves a markedly lower reconstruction error (MAE~0.0046) than the standard CNN-based VQGAN (MAE~0.0134), while using fewer parameters (43M vs.\ 104M).
This improvement arises from replacing the convolutional encoder–decoder with a Vision Transformer (ViT) backbone and adopting a larger latent grid ($25 \times 25 \times 196$) without a compression bottleneck.
The expanded latent representation preserves fine geological structures that are often lost when previous models heavily downsample the latent space.
These results confirm that this new ViT-VQGAN tokenizer offers superior reconstruction fidelity, providing a more expressive and physically meaningful latent representation. 

\begin{table}[h]
\centering
\small
\vspace{-2mm}
\caption{\textbf{Quantitative comparison} of VQGANs.}
\vspace{-2mm}
\label{tab:vqgan}
\begin{tabular}{lccc}
\toprule
Model & MAE $\downarrow$ & Dim & \#Parameters \\
\midrule
VQGAN & 0.0134 & 9x9x32 & 104M \\
ViT-VQGAN & 0.0046 & 25x25x196 & 43M \\
\bottomrule
\end{tabular}
\vspace{-4mm}
\end{table}

\section{Ablation study.} \label{app:ablation}
We conduct additional controlled ablations to disentangle the effects of model capacity and data scale. All models are evaluated without RL or latent GD to ensure a fair comparison. The \emph{Small} and \emph{Shallow} variants are designed to have middle-sized parameter counts (approximately 300M–400M). Specifically, \emph{Small} reduces the width by half while increasing the depth from 16 to 24 layers, whereas \emph{Shallow} reduces the depth from 16 to 6 layers.

Table~\ref{tab:ablation} shows three key observations. 
First, \textbf{data scaling primarily improves performance}: training our model on Data $\times 2$ reduces MAE from 0.0227 to 0.0167, indicating that increased data diversity significantly benefits reconstruction accuracy. 
Second, \textbf{model capacity plays an important role}: under the same Data $\times 2$ setting, smaller variants consistently underperform the 1B model (0.0195/0.0217 vs. 0.0167), demonstrating that the gains are not solely due to data scaling. 
Third, \textbf{architecture matters beyond parameter count}: BigFWI-L (87M) performs substantially worse (0.0397) even with matched data scale, suggesting that the proposed transformer-based design is more effective for this task.

Overall, these results confirm that \textbf{data scale, model capacity, and architectural design each contribute meaningfully}, and the observed improvements cannot be attributed to a single factor.

\begin{table}[h]
\small
\centering
\caption{\textbf{Ablation study.}}
\vspace{-2mm}
\label{tab:ablation}
\begin{tabular}{llll}
\toprule
Model  & Data  & OpenFWI MAE & \#Param \\ \hline
Ours         & Data x2 & \bf{0.0167 }         & 1.03B        \\ \hline
Shallow     & Data x2 & 0.0217            & 422M         \\ \hline
Small  & Data x2 & 0.0195          & 313M         \\ \hline
BigFWI-L     & Data x2 & 0.0397            & 87M     \\    \hline
Ours         & OpenFWI & 0.0227           & 1.03B        \\ \hline
\end{tabular}
\vspace{-4.5mm}
\end{table}




\end{document}